\documentclass{article}

\usepackage{arxiv}
\usepackage{graphicx}
\graphicspath{{Figures/},.}
\usepackage{amsfonts, bm, amsmath, amssymb}
\usepackage{mathtools}
\usepackage[table]{xcolor}
\usepackage{xcolor}
\usepackage{tabularx}
\usepackage{pdflscape}
\usepackage{multirow}

\definecolor{myred}{RGB}{153, 0, 0}
\definecolor{highlight}{RGB}{254, 232, 200}

\usepackage{makecell}
\usepackage{setspace}

\usepackage{comment}
\usepackage[utf8]{inputenc} 
\usepackage[T1]{fontenc}    
\usepackage{hyperref}       
\usepackage{url}            
\usepackage{booktabs}       
\usepackage{nicefrac}       
\usepackage{microtype}      
\usepackage[numbers,sort&compress]{natbib}

\usepackage{doi}
\usepackage[ruled]{algorithm2e}
\usepackage{algpseudocode}
\newcommand*{\vertbar}{\rule[-1ex]{0.5pt}{2.5ex}}
\newcommand*{\etal}{\textit{et al.}}
\newcommand*{\fref}{Fig.~\ref}

\algrenewcommand\algorithmicrequire{\textbf{Input:}}
\algrenewcommand\algorithmicensure{\textbf{Output:}}
\algnewcommand{\LeftComment}[1]{\Statex \(\triangleright\) #1}

\title{Automating the Discovery of Partial Differential Equations in Dynamical Systems}
\date{} 					

\author{ 
	\href{https://orcid.org/0000-0001-9368-236X}{\includegraphics[scale=0.06]{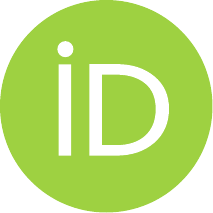}\hspace{1mm}Weizhen Li} \\
	Department of Engineering\\
        Durham University\\
	Durham, UK, DH1 3HN\\
	\texttt{weizhen.li@dur.ac.uk} \\
    \And
    \href{https://orcid.org/0000-0002-3279-4218}{\includegraphics[scale=0.06]{orcid.pdf}\hspace{1mm}Rui Carvalho} \\
	Department of Engineering\\
        Durham University\\
	Durham, UK, DH1 3HN\\
	\texttt{rui.carvalho@dur.ac.uk} \\
}

\renewcommand{\shorttitle}{\textit{arXiv} Template}
\hypersetup{
pdftitle={A template for the arxiv style},
pdfsubject={q-bio.NC, q-bio.QM},
pdfauthor={David S.~Hippocampus, Elias D.~Striatum},
pdfkeywords={First keyword, Second keyword, More},
}

\begin{document}
\maketitle

\begin{abstract}
Identifying partial differential equations (PDEs) from data is crucial for understanding the governing mechanisms of natural phenomena, yet it remains a challenging task. We present an extension to the ARGOS framework, ARGOS-RAL, which leverages sparse regression with the recurrent adaptive lasso to identify PDEs from limited prior knowledge automatically. Our method automates calculating partial derivatives, constructing a candidate library, and estimating a sparse model. We rigorously evaluate the performance of ARGOS-RAL in identifying canonical PDEs under various noise levels and sample sizes, demonstrating its robustness in handling noisy and non-uniformly distributed data. We also test the algorithm's performance on datasets consisting solely of random noise to simulate scenarios with severely compromised data quality. Our results show that ARGOS-RAL effectively and reliably identifies the underlying PDEs from data, outperforming the sequential threshold ridge regression method in most cases. We highlight the potential of combining statistical methods, machine learning, and dynamical systems theory to automatically discover governing equations from collected data, streamlining the scientific modeling process.
\end{abstract}


\section{Introduction}

In recent years, scientists have increasingly employed statistical and machine learning methods to uncover the governing equations of dynamical systems, particularly differential equations, from observational data~\cite{schmidt_distilling_2009,xun_parameter_2013,brunton_discovering_2016,rudy_data-driven_2017,raissi_physics-informed_2019}. 
Data-driven methods offer several advantages over traditional approaches that rely on first principles and expert knowledge. 
These methods can reveal patterns and relationships in the data that may not be apparent from first principles, providing new insights into complex systems~\cite{lu_discovering_2022,zhang_analyses_2022}. 
They are also adept at working with noisy or incomplete data commonly encountered in real-world applications, employing techniques from machine learning to enhance the robustness of discoveries~\cite{jiang_modeling_2021,maddu_stability_2022,cai_online_2023,sun_compressive-sensing_2024}. 
Furthermore, by reducing the need for manual intervention and domain expertise, data-driven methods can significantly streamline the discovery process~\cite{egan_automatically_2024}.

Data-driven discovery in dynamical systems has evolved from early parameter estimation using spline approximation and system reconstruction~\cite{varah_spline_1982,crutchfield_equations_1987}, to leveraging statistical methods such as least squares~\cite{bar_fitting_1999,muller_fitting_2002,liang_parameter_2008}, mixed-effects models~\cite{wu_estimation_1998,wu_population_1999}, and Bayesian approaches~\cite{putter_bayesian_2002,xun_parameter_2013} for parameter estimation in ordinary and partial differential equations (ODEs and PDEs). 
The advent of high-performance computing has further propelled symbolic regression, enabling the discovery of governing equations from data in physics and engineering~\cite{bongard_automated_2007,schmidt_distilling_2009,udrescu_ai_2020,xu_robust_2021}. 
A notable development in this field is the Sparse Identification of Nonlinear Dynamics (SINDy) approach~\cite{brunton_discovering_2016,rudy_data-driven_2017}, which constructs an extensive library of potential terms and employs the Sequential Threshold Ridge Regression (STRidge) algorithm~\cite{rudy_data-driven_2017} to select significant terms iteratively. 

SINDy and its various enhancements~\cite{rudy_data-driven_2019,kaheman_sindy-pi_2020,messenger_weak_2021,cortiella_sparse_2021,fasel_ensemble-sindy_2022,li_discover_2023} have been extensively used to discover a broad spectrum of ODEs and PDEs, describing diverse phenomena such as fluid mechanics~\cite{loiseau_sparse_2018}, turbulence models~\cite{duraisamy_turbulence_2019}, aerodynamics~\cite{li_discovering_2019}, and biological and chemical systems~\cite{mangan_inferring_2016,hoffmann_reactive_2019}. 
Recent developments have combined neural network-based techniques and SINDy, leading to innovative approaches that enhance noise tolerance in identifying PDEs~\cite{raissi_physics-informed_2019,lagergren_learning_2020,xu_robust_2021,chen_physics-informed_2021,zhang_robust_2021,thanasutives_noise-aware_2023,jia_governing_2023}. 
Neural networks can learn complex nonlinear relationships and effectively filter out noise, complementing SINDy's ability to identify parsimonious models. 
However, both neural network and SINDy methods require specific hyperparameter tuning, such as setting regularization parameters or choosing network architectures.
For example, STRidge requires setting a threshold to select active terms from the candidate library~\cite{rudy_data-driven_2017,chen_physics-informed_2021,thanasutives_noise-aware_2023,li_discover_2023}.
Additionally, SINDy-based methods typically approximate numerical derivatives from noisy data using the Savitzky-Golay filter, a technique for smoothing data by fitting local low-degree polynomials~\cite{savitzky_smoothing_1964}. 
The parameters of this filter, such as the polynomial degree and window size, must be carefully tuned for optimal performance~\cite{rudy_data-driven_2017,egan_automatically_2024}.
Neural network approaches, on the other hand, require detailed decisions regarding their architecture and functioning, such as the number of neurons, the structure of hidden layers, the types of activation and loss functions, and the learning rate.
In particular, using physics-informed neural networks~\cite{raissi_physics-informed_2019,chen_physics-informed_2021,thanasutives_noise-aware_2023} requires a prior understanding of the equation terms, as well as initial and boundary conditions.
Consequently, using neural networks and SINDy-based methods presents a trade-off: the absence of fully automated algorithms requires users to engage in manual tuning and iterative usage of semi-automated algorithms.
This scenario highlights a key challenge in the field: developing an automated algorithm to identify PDEs with minimal manual intervention, streamlining the process, and improving its applicability across diverse scientific domains.

To address the challenge of parameter tuning, Egan \etal~\cite{egan_automatically_2024} proposed the Automatic Regression for Governing Equations (ARGOS) algorithm, which identifies ODEs by automating the parameter tuning process.
ARGOS assumes the underlying system is unknown, automates the fine-tuning of parameters for numerical differentiation, and leverages sparse regression with bootstrap confidence intervals to select active terms from the candidate library. 
To automatically identify PDEs, we develop ARGOS with the Recurrent Adaptive Lasso (ARGOS-RAL). 
This extension of the ARGOS framework employs only sparse regression to identify equations rather than engaging in large-scale bootstrapping.

We evaluate the performance of the ARGOS-RAL algorithm through a series of three numerical tests, each designed to assess its ability to identify canonical PDEs across diverse fields, including biology, neuroscience, earth science, fluid mechanics, and quantum mechanics.
The first test explores the algorithm's resilience against varying noise levels by altering the signal-to-noise ratio (SNR) in Gaussian random noise integrated into the PDE solutions, which is crucial for understanding the robustness of ARGOS-RAL under realistic noisy conditions.
The second test addresses the practical challenges encountered in real-world data collection, which often results in non-uniformly distributed data points in space and time, by exploring the minimum percentage of data points necessary for the algorithm to accurately identify the underlying equation.
The final evaluation assesses the algorithm's ability to process datasets characterized by significant noise, challenging its limits and practical applicability in scenarios where data quality is compromised. 
Our results demonstrate that ARGOS-RAL can effectively and reliably identify the underlying PDEs from data, outperforming the STRidge method used in SINDy.

\section{Methods}

\subsection{Overview of the ARGOS-RAL Framework}
The general form of a homogeneous PDE is
\begin{equation}\label{eq:pdege0}
    u_t + F (x, t, u, u_x , u_{xx}, \cdots) = 0
\end{equation}
where $F(\cdot)$ governs the behavior of the system, with $u=u(x,t)$ denoting its state. The notation $u_t,u_x,u_{xx}, \cdots$ represents the partial derivatives of $u$ with respect to time and space, respectively. 
Equation~\eqref{eq:pdege0} serves as a foundational representation of the dynamical system, encapsulating a wide range of phenomena through its generalized form, which can be adapted to include multiple spatial dimensions or to model systems without explicit time dependence.

To focus on data-driven modeling of spatiotemporal dynamical systems, we incorporate empirical data directly into the modeling process:
\begin{equation} \label{eq:pdege01}
   \mathbf{U}_t = \frac{\partial \mathbf{U}}{\partial t} = \mathbf{F}\left(x,\mathbf{U},\mathbf{U}_x,\mathbf{U}_{xx},\dots\right),
\end{equation}
where $\mathbf{U} \in \mathbb{R}^{n \times m}$ is a matrix representing the solution of the PDE as a function of $x$ and $t$, and $\mathbf{F}(\cdot)$ denotes the unknown mapping inferred from the collected data, which contains linear and nonlinear operators. 

We aim to estimate the unknown mapping $\mathbf{F}(\cdot)$ with sparse regression by constructing a comprehensive library of potential terms and assuming that only a few of them are active~\cite{brunton_discovering_2016,rudy_data-driven_2017,egan_automatically_2024}. 
To cover a broad spectrum of possible influences on the dynamics of the system, this library includes a wide variety of functions, such as constants, monomials, interaction terms (products of variables), possibly trigonometric, and other functions, depending on the dynamical system being studied~\cite{rudy_data-driven_2017}. 
In the case of Burgers' equation, $u_t = -uu_x + 0.1 u_{xx}$, the true dynamics involves only two terms: the nonlinear convection term $uu_x$ and the linear diffusion term $u_{xx}$. When applying sparse regression to data generated from Burgers' equation, the method should ideally select only these two terms from the candidate library. 

All features related to $\mathbf{U}(x,t)$ in Eq.~\eqref{eq:pdege01} are matrices.
Implementing this matrix data in sparse regression leads to the creation of $m$ distinct regression models. 
Each model captures the spatial dynamics of the system at a specific time point $t_j$, where $j=1,2,\dots,m$. 
To consolidate the $m$ regression models into a single linear regression problem, we reshape the matrix $\mathbf{U}(x,t)$ and its derivative matrices into vectors. 
These vectors then serve as predictors within the candidate library $\mathbf{\Theta}$, which can be represented in $\mathbb{R}^{(n\cdot m) \times p}$ or $\mathbb{C}^{(n\cdot m) \times p}$. 
By stacking the vectorized data and candidate terms, we can estimate a single sparse coefficient vector $\beta$ that represents the governing equation across all time points rather than estimating separate models for each time point. 
Here, $\mathbf{U}\in \mathbb{R}^{n\times m}$ is represented in matrix form as
\begin{equation} \label{eq:u_pde}
\mathbf{U}(x,t)
    = \left(\begin{array}{cccc}
    u(x_1,t_{1}) & u(x_1,t_{2}) & \cdots & u(x_1,t_{m}) \\
    u(x_2,t_{1}) & u(x_2,t_{2}) & \cdots & u(x_2,t_{m}) \\
    \vdots  & \vdots  & \ddots & \vdots  \\
    u(x_n,t_{1}) & u(x_n,t_{2}) & \cdots & u(x_n,t_{m})
    \end{array}\right).
\end{equation}
Vectorizing Eq.~\eqref{eq:u_pde} yields:
\begin{equation} \label{eq:u_pde2}
    \mathbf{u} = vec(\mathbf{U}) =
        \left(\begin{array}{ccccccc}
    u(x_1,t_{1}) & \cdots & u(x_n,t_{1}) & \cdots &  u(x_1,t_{m}) & \cdots & u(x_n,t_{m})
    \end{array}\right) ^T .
\end{equation}
Similarly,  
$\mathbf{u}_t = vec(\mathbf{U}_t) = vec({\partial \mathbf{U}}/{\partial t})$, $\mathbf{u}_x = vec(\mathbf{U}_x) = vec({\partial \mathbf{U}}/{\partial x})$, $\mathbf{u}_{xx} = vec(\mathbf{U}_{xx}) = vec({\partial^2 \mathbf{U}}/{\partial x^2})$, $\mathbf{u}^2 = vec(\mathbf{U}\odot\mathbf{U})$, and $\mathbf{uu}_x = vec(\mathbf{U} \odot \mathbf{U}_x)$, where $\odot$ denotes the Hadamard product. 
The design matrix is given by
\begin{equation} \label{eq:theta_pde2} 
\hspace{-10pt}
    \mathbf{\Theta}(\mathbf{u}) = 
    \left(\begin{array}{cccccccccccc}
        \vertbar & \vertbar & & \vertbar &  & \vertbar & \vertbar & & \vertbar &  & \vertbar &  \\
        \mathbf{1}  & \mathbf{u} & \cdots & \mathbf{u}^{d} & \cdots  & \mathbf{u}_{x} & \mathbf{u}_{xx} & \cdots & \mathbf{uu}_{x} & \cdots & \mathbf{u}^d \mathbf{u}_{xx} & \cdots \\
        \vertbar & \vertbar & & \vertbar &  & \vertbar & \vertbar & & \vertbar &  & \vertbar &
    \end{array}\right),
\end{equation}
where $\mathbf{u}^{d}$ is a vector where all elements denote a $d$-th degree monomial.
For example, if our data $\mathbf{U}(x,t)$ is on a $200 \times 100$ grid (i.e. 200 spatial measurements and 100 time-steps) and the candidate library has 30 terms, then $\mathbf{\Theta} \in \mathbb{R}^{20000 \times 30}$. 

After vectorization, we estimate $\mathbf{F}(\cdot)$ by transforming Eq.~\eqref{eq:pdege01} to a linear regression model
\begin{equation}\label{eq:pde_linear}
      \mathbf{u}_t =  \mathbf{\Theta}(\mathbf{u})\beta+\epsilon, 
\end{equation}
where $\beta \in \mathbb{R}^{p}$ is a sparse coefficient vector in which only a few values are nonzero, and $\epsilon$ is the vector of residuals. 

\begin{figure}[htp]
    \centering
    \includegraphics[width=1\textwidth]{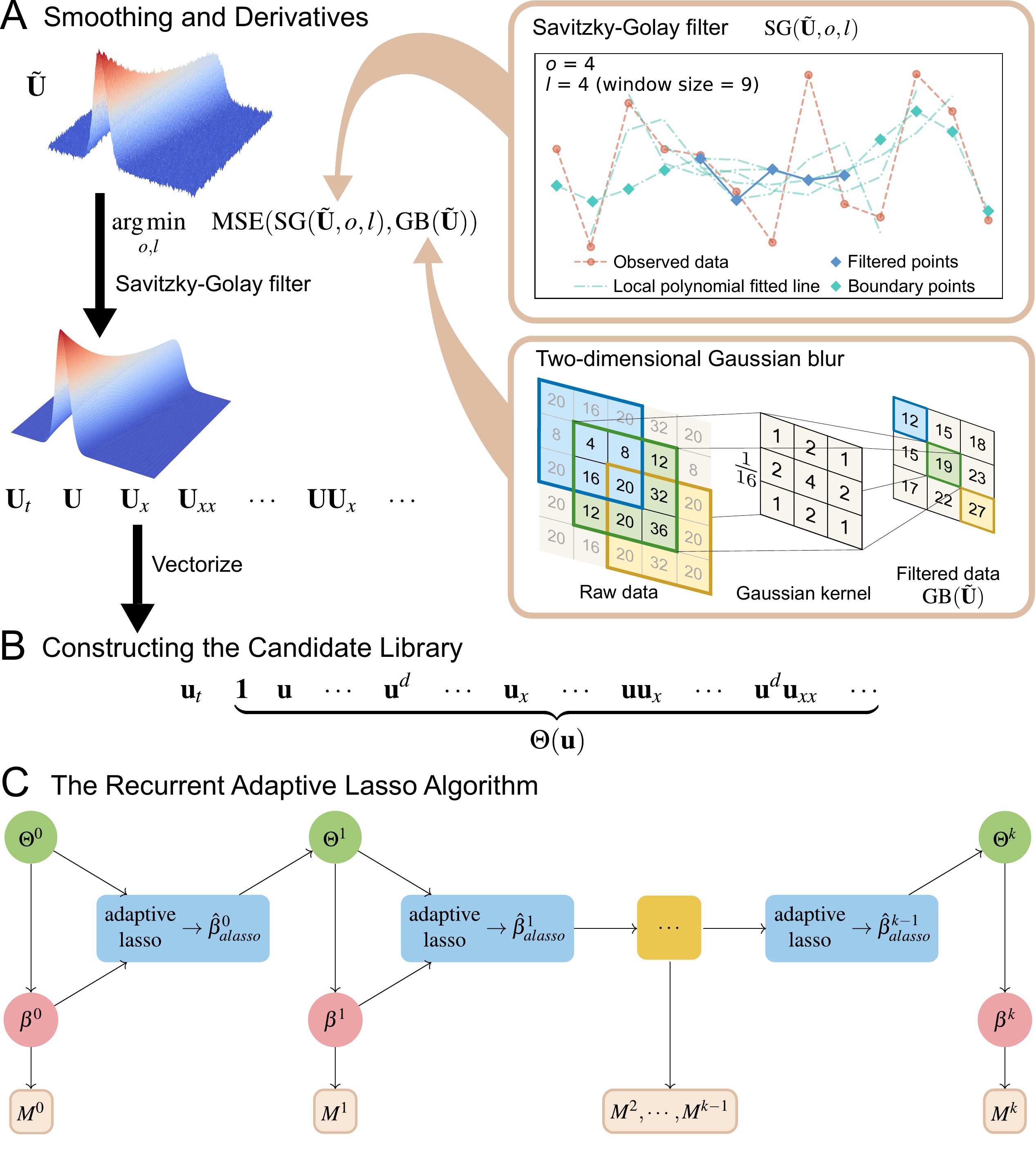}
    \caption{Process of identifying PDEs from data using ARGOS with the recurrent adaptive lasso. 
    The identification process consists of three main steps: (A) automatic smoothing and calculation of derivatives, (B) construction of the candidate library, and (C) implementation of the recurrent adaptive lasso. 
    We begin by collecting the data $\tilde{\mathbf{U}}$ and applying the automatic Savitzky-Golay filter with Gaussian blur to calculate the smoothed $\mathbf{U}$ and its partial derivatives. Next, we vectorize the smoothed data, all partial derivatives, and other related terms to construct the candidate library.
    Finally, we employ the recurrent adaptive lasso to identify the active features in the library, and we estimate the unbiased coefficients of the identified model using ordinary least squares regression.}
    \label{dia:id_pde}
\end{figure}

\subsection{Automated Numerical Differentiation using the Savitzky-Golay Filter and the Gaussian Blur} \label{sec:ASGGB}

A crucial step in constructing the candidate library in Eq.~\eqref{eq:theta_pde2} is the numerical calculation of derivatives (see \fref{dia:id_pde} A and B). The Savitzky-Golay filter~\cite{savitzky_smoothing_1964} has become a favored solution in system identification for signal smoothing and differentiation~\cite{rudy_data-driven_2017,breugel_numerical_2020}.
This method applies a least squares polynomial fit over a sliding window of data points, thereby achieving simultaneous signal smoothing and differentiation. 
The selection of the Savitzky-Golay filter is grounded in its proven ability to accurately maintain the original contour of the signal while significantly reducing noise and to approximate higher-order numerical derivatives with symbolic differentiation~\cite{schafer_what_2011}. 

The Savitzky-Golay filter is characterized by two integer hyperparameters: the polynomial order $o$ and the window length $l$, which are constrained by the conditions that $o$ must be at least 2, $l$ should be an odd number, and $o+1+mod(o)\leq l \leq n-1$~\cite{schafer_what_2011}. 
To automate the selection of these hyperparameters, we first apply a Gaussian blur with the kernel $(1,2,1)$ to smooth the observational data (see \ref{app:GB}).
We then treat this smoothed data, denoted as GB$(\tilde{\mathbf{U}})$, as the ground truth.
Next, we find the optimal set of hyperparameters $\{o^\ast,l^\ast\}$ by minimizing the mean squared error (MSE) between the Savitzky-Golay filtered data SG$(\tilde{\mathbf{U}},o,l)$ and the ground truth GB$(\tilde{\mathbf{U}})$ (see Algorithm~\ref{al:SG} in \ref{app:algos}). 
After finding the optimal set $\{o^\ast,l^\ast\}$, we use the Savitzky-Golay filter with these parameters to compute the smoothed data and its derivatives.

\subsection{Sparse Regression with the Recurrent Adaptive Lasso} \label{sec:RAL}

The adaptive lasso is a two-step method~\cite{zou_adaptive_2006,egan_automatically_2024}. 
The first step uses the ordinary least squares (OLS) to obtain unbiased estimates and derive the weights $w$: 
\begin{equation} \label{eq:adalasso2} 
    w = |\hat{\beta}_{ols}|^{-\gamma}, \quad \gamma>0 
\end{equation}
where $\hat{\beta}_{ols}$ is the OLS estimate, and $\gamma$ is an exponent tuning the shape of the soft-thresholding function.
In the second step, we obtain the estimated coefficients $\hat{\beta}_{alasso}$ using the glmnet package~\cite{friedman_regularization_2010} in R by solving the problem
\begin{equation} \label{eq:adalasso} 
    \hat{\beta}_{alasso}=\mathop{\arg\min}_{\beta} \left\|u_t - \mathbf{\Theta}\beta\right\|^2_2 + \lambda\sum^{p}_{j=1}w_j\left|\beta_j\right| , 
\end{equation}
where $\lambda$ is a nonnegative regularization parameter controlling the amount of shrinkage applied to the coefficients of the predictors. 
Unlike the lasso, where the weight vector is $w=\mathbf{1}$, the adaptive lasso varies the weights in the regularization function, resulting in a stronger penalty on smaller coefficients, thus driving more of them to zero and leading to a sparser model compared to the standard lasso. 
The recurrent adaptive lasso applies the adaptive lasso repeatedly until convergence, resulting in a sparse model with fewer non-zero coefficients. 

\begin{figure}[htp]
    \centering
    \includegraphics[width=0.7\linewidth]{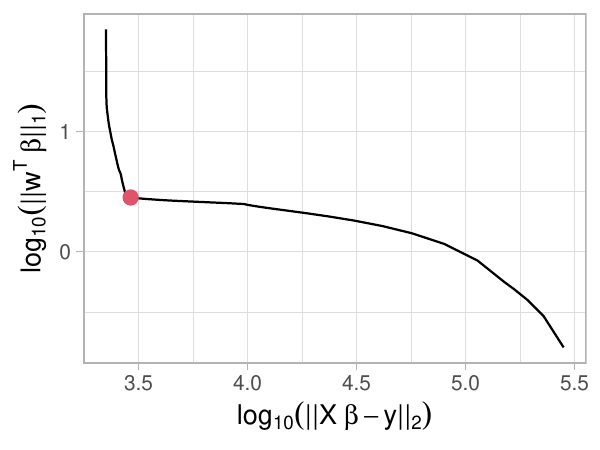}
    \caption{Pareto curve of the adaptive lasso for a sampled dataset from a Navier-Stokes system with an SNR of 36 dB. The Pareto curve balances the trade-off between sparsity and goodness-of-fit. The red point on the curve indicates the optimal value of the regularization parameter $\lambda$ that achieves the best balance between these two competing objectives. Increasing $\lambda$ leads to sparser solutions at the cost of a poorer fit to the data, while decreasing $\lambda$ improves the fit but yields less sparse solutions. }
    \label{fig:pareto_curve}
\end{figure}

To balance the model's complexity against its accuracy, we determine the regularization parameter $\lambda$ by employing the Pareto curve, which illustrates the optimal trade-off between the regularization penalty and the model residuals~\cite{hansen_analysis_1992,nasehi_tehrani_l1_2012,cultrera_simple_2020} (see \fref{fig:pareto_curve}).
Although cross-validation is an alternative method, Cortiella \etal~\cite{cortiella_sparse_2021} have shown that it finds a $\lambda$ optimized for prediction, potentially overfitting the true underlying equation with extra features.

The adaptive lasso regression often detects more terms than those in the true system. 
To improve parsimony, Egan \etal~\cite{egan_automatically_2024} suggested combining the adaptive lasso with bootstrap techniques to identify ODEs. 
Similarly, Cortiella \etal~\cite{cortiella_sparse_2021} adopted a modified version of the multi-step adaptive lasso~\cite{buhlmann_statistics_2011} to develop a sparser model that more accurately identifies the true equations. 
This is achieved by iteratively adjusting the adaptive weights using previous estimates from the adaptive lasso. 
A significant advancement made by Cortiella \etal~\cite{cortiella_sparse_2021} is their method's ability to maintain finite weights in the adaptive lasso equation by ensuring that the estimated coefficients shrink to a small, nonzero value rather than dropping to zero. 
However, this approximation unintentionally introduces numerical inaccuracies as a trade-off for preventing overflow during the equation identification process.

The recurrent adaptive lasso is an iterative algorithm that estimates an initial sparse model using the adaptive lasso and subsequently refining it by trimming the candidate library (see \fref{dia:id_pde} C).  
At each iteration, it removes terms whose coefficients the adaptive lasso penalized to zero (see \ref{app:algos} Algorithm~\ref{al:reada} step 9). It then employs least squares to re-estimate the coefficients of the remaining terms, which are used to update the adaptive weights in the next adaptive lasso iteration. This focuses the regularization on the terms that had small coefficients in the previous iteration.
As this process repeats, the recurrent adaptive lasso increasingly concentrates the $\ell_1$-norm shrinkage on terms that are likely irrelevant, driving their coefficients to zero~\cite{tibshirani_regression_1996,zou_adaptive_2006}. Meanwhile, it relaxes the regularization on terms that consistently have larger coefficients, allowing the model to retain them. The candidate set gets smaller at each iteration until the algorithm converges on a sparse model containing only the key terms.
This iterative re-weighting allows the recurrent adaptive lasso to prune irrelevant terms more aggressively than the standard adaptive lasso while retaining good predictive performance. The result is a parsimonious model that identifies the true governing equation more reliably, even in the presence of many extraneous candidate terms.

Increasing the number of iterations may cause the recurrent adaptive lasso to underestimate the model. This can lead to the omission of active terms that should be included in the true underlying equation.
Therefore, while iterating the candidate library $\mathbf{\Theta}$, we record all candidate models and calculate the Akaike information criterion (AIC) for each model to determine the final governing equation corresponding to the lowest AIC.
Given the uncertainty that the true model falls within all candidates, the AIC serves to select the model that best approximates the true model~\cite{yang_can_2005,aho_model_2014}.

\section{Results and Discussion}

\subsection{Evaluating the Performance of ARGOS-RAL under Varying Noise Levels and Sample Sizes}
\begin{figure}[htp]
    \centering
    \includegraphics[width=\textwidth]{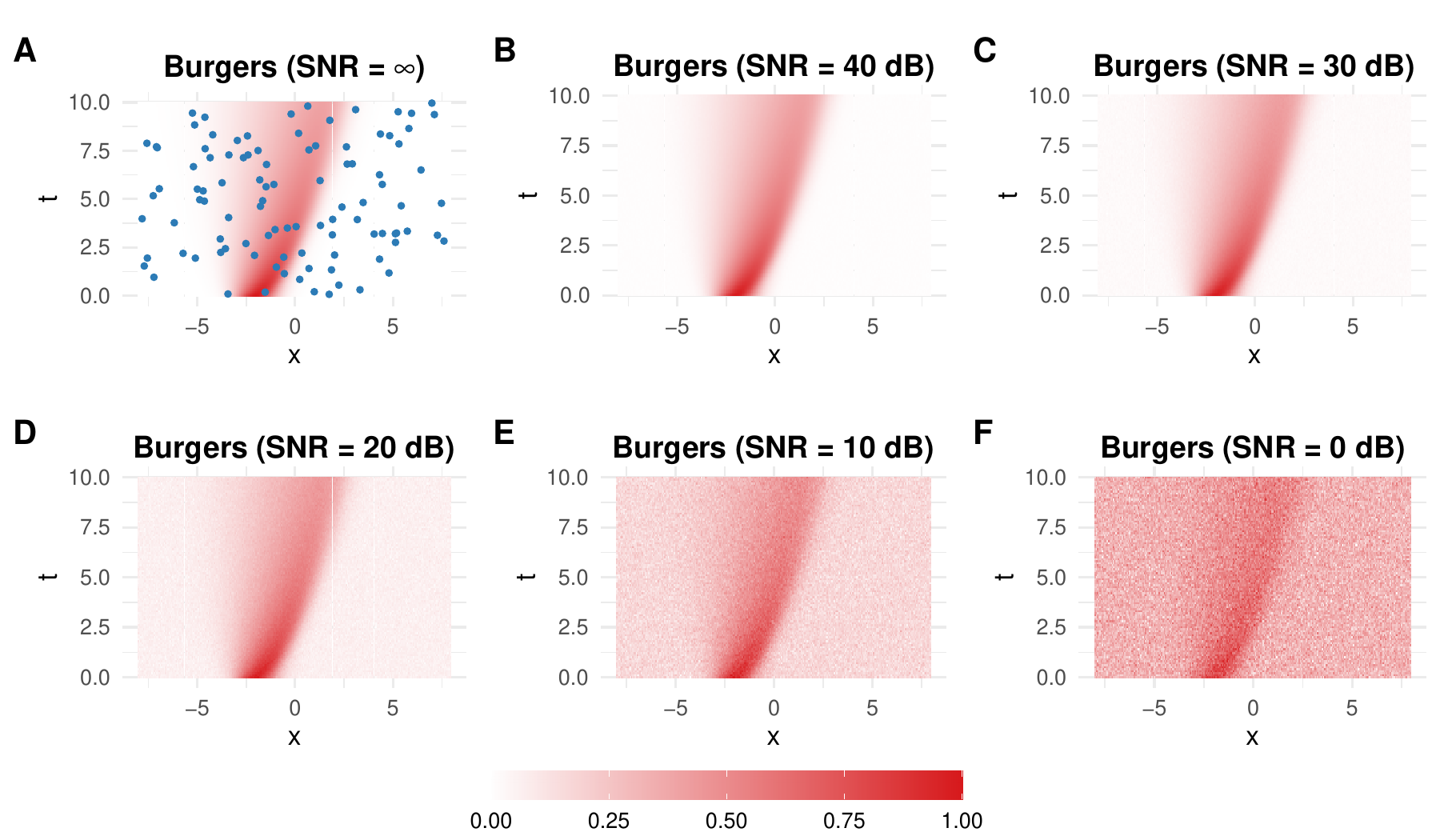}
    \caption{Influence of SNR on the Burgers' equation dataset. (A) Noiseless data points (blue) serve as a reference for evaluating the impact of sample size on PDE identification accuracy. (B-F) Noisy datasets are generated by adding Gaussian noise at SNR levels of 40 dB, 30 dB, 20 dB, 10 dB and 0 dB, respectively, to comprehensively characterize the system's behavior under varying noise conditions.}
    \label{fig:bur_noise}
\end{figure}

We compare the performance of ARGOS-RAL and STRidge~\cite{rudy_data-driven_2017} in identifying ten canonical PDEs under various SNRs and sample sizes ($N$). We evaluate their performance on both noisy and noiseless data.
Figure~\ref{fig:bur_noise} demonstrates the impact of introducing increasing levels of Gaussian random noise into the solution of the Burgers' equation, effectively decreasing the SNR values. 

In the evaluation of noise-contaminated data, we express the SNR as $\mathrm{SNR}=20\log_{10}(\sigma_{\mathbf{U}} / \sigma_{Z})$, where $\sigma_{\mathbf{U}}$ is the standard deviation of the original data, and $\sigma_{Z}$ represents the standard deviation of the added noise.
We systematically vary $\sigma_{Z}$ to span a broad range of noise levels, facilitating a comprehensive evaluation of the efficacy of ARGOS-RAL and STRidge in identifying various PDEs under different noise conditions. 
For this purpose, we generate datasets with SNRs set at $\{0, 2, \cdots, 58, 60, \infty\}$~\cite{egan_automatically_2024}, each comprising paired elements $\left\{\mathbf{u}_t, \mathbf{\Theta}(\mathbf{u})\right\}$. 
This approach allows us to examine the robustness of each PDE identification method as it copes with varying noise levels.

In investigating sample size, $N$, our objective is to determine the smallest number of samples needed to reliably identify PDEs with a success rate exceeding 80\%.
To achieve this, we first generate a full dataset for each PDE by calculating partial derivatives and assembling a candidate library as described in Eq.~\eqref{eq:theta_pde2}. The size of the full dataset, denoted as $\mathbf{N}$, varies depending on the specific PDE under consideration.
Specifically, $\mathbf{N}=10^4$ for the advection-diffusion, Burgers, and cable equations, $\mathbf{N}=10^5$ for the quantum harmonic oscillator, transport, Navier-Stokes, and reaction-diffusion equations, and $\mathbf{N}=10^{4.8}$ for the heat and Korteweg-De Vries (KdV) equations.
Next, we randomly sample smaller subsets of size $N$ from the full dataset, where $N$ is chosen from a logarithmically spaced grid: $N=10^2, 10^{2.2}, 10^{2.4}, \cdots, \mathbf{N}$~\cite{egan_automatically_2024} (see the blue points in \fref{fig:bur_noise} A).
By applying the PDE identification methods to these subsets and evaluating their success rates, we can determine the smallest sample size required for reliable identification of each PDE. 

\subsection{Quantifying Success Rates in Identifying Canonical PDEs}

\begin{figure}[htp]
    \centering
    \includegraphics[width=1\textwidth]{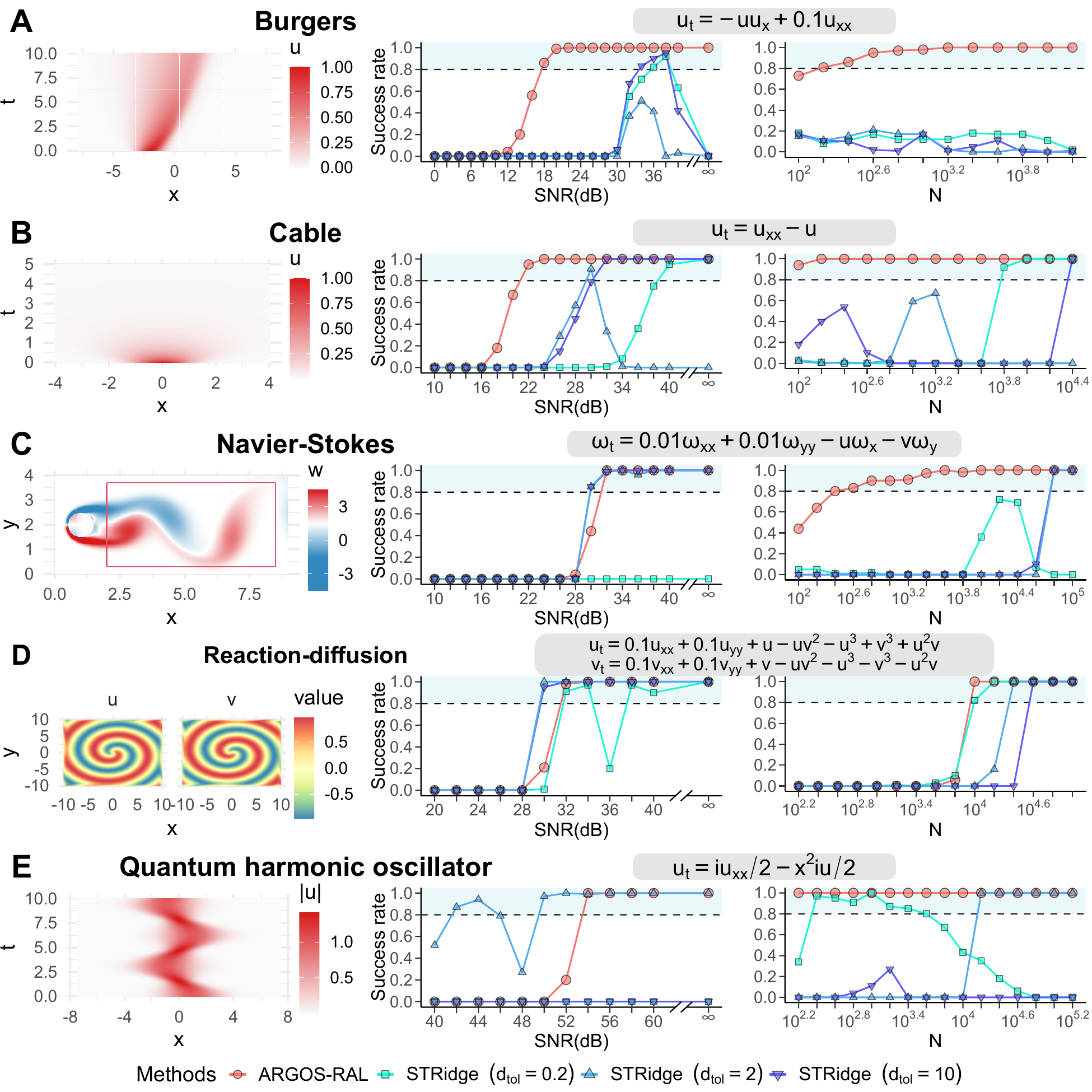}
    \caption{Success rates of ARGOS-RAL and STRidge in identifying (A) Burgers', (B) cable, (C) Navier-Stokes, (D) reaction-diffusion, and (E) quantum harmonic oscillator equations with varying SNRs and sample sizes. We analyze the noise tolerance by adding noise of different SNRs to the PDE solutions. For the sample size analysis, we randomly sample points from the set $\{\mathbf{u}_t,\mathbf{\Theta}(\mathbf{u})\}$ based on noiseless data. In panel (C), we use the region indicated by the red rectangle to implement both the SNR and sample size tests by sampling points within this area. PDE solution plots display time snapshots at $t=306$ for Navier-Stokes in panel (C) and $t=1$ for reaction-diffusion in panel (D). Lines connecting the points are used for visual guidance only and do not represent a fit to the data. Shaded regions represent model discovery accuracy above 80\%. 
    }
    \label{fig:all_plots}
\end{figure}

To evaluate the impact of different SNRs and data sizes on the method, we measure the uncertainty of model identification caused by random sampling.
To do so, we create 100 unique datasets at each point on the grid, corresponding to different SNRs and $N$ values.
For each dataset, we quantify the identification accuracy with the success rate, $\eta = \mathrm{\#correct}/100$, where $\mathrm{\#correct}$ represents the number of times the model correctly identifies all active terms. 
Our accuracy assessment ignores small differences between theoretical and empirical coefficients, such as a theoretical value of 0.1 compared to an estimated value of 0.098. 
Figure~\ref{fig:all_plots} illustrates these results for a selection of systems: the Burgers', Cable, Navier-Stokes, reaction-diffusion, and quantum harmonic oscillator models.
We provide further analysis on additional PDEs --Transport, Heat, Advection-Diffusion, and KdV equations -- in \ref{app:more_eq}~\fref{fig:all_plots2}.

ARGOS-RAL identifies Burgers', cable, Navier-Stokes, reaction-diffusion, and advection-diffusion equations, achieving a success rate of 100\% when the SNR exceeds 30 dB (see \fref{fig:all_plots} and \ref{fig:all_plots2}).
However, accurately detecting specific equations requires a high SNR, particularly for the quantum harmonic oscillator, KdV, transport, and diffusion equations.
The KdV equation, which involves third-order partial derivatives, presents challenges due to the significant biases in numerical approximations of these derivatives~\cite{jia_governing_2023}, resulting in datasets unsuitable for system identification with sparse regression.
To implement sparse regression within the real number domain for the complex number quantum harmonic oscillator PDE, we apply the transformation shown in \ref{app:more_eq} Eq.~\eqref{eq:complex}.
This transformation expands the design matrix $\mathbf{\Theta}$ from $nm \times p$ to $2nm \times 2p$, effectively quadrupling its size and potentially leading to high correlations between the variables in $\mathbf{\Theta}$.
The transport and diffusion equations, containing only terms $u_x$ and $u_{xx}$ respectively, exhibit high correlation with their correlated terms in the library, such as $\{u_x, uu_x\}$ and $\{u_{xx}, uu_{xx}\}$, which hinders the effectiveness of $\ell_1$-norm shrinkage regression in identifying correct terms~\cite{zou_adaptive_2006}.

Figures \ref{fig:all_plots} and \ref{fig:all_plots2} illustrate that ARGOS-RAL achieves a higher success rate than STRidge in identifying PDEs with limited data points.
ARGOS-RAL consistently identifies a significant number of PDEs using as few as 1000 data points, maintaining a success rate above 80\%. 
However, some equations, such as the reaction-diffusion and KdV equations, require larger sample sizes of approximately $10^4$ and $10^{3.8}$ data points, respectively, for reliable identification.
We thus demonstrate ARGOS-RAL as a consistent and efficient method for PDE identification with non-uniformly sampled and noiseless datasets.

ARGOS-RAL shows a remarkable ability to identify PDEs accurately and consistently across a wide range of SNRs and sample sizes. 
Its success rate improves as the SNR and sample size increase, reaching 100\% when both values are sufficiently large. This trend highlights the robustness of ARGOS-RAL in handling various data conditions and underscores its effectiveness in identifying PDEs, even when faced with varying levels of data quality and quantity. 
However, in certain scenarios, STRidge~\cite{rudy_data-driven_2017} with specific $d_{tol}$ thresholds exceeds the performance of ARGOS-RAL. 
For instance, STRidge achieves higher success rates in identifying Navier-Stokes and reaction-diffusion equations at a 30 dB SNR, using $d_{tol}$ settings of 2 and 10, respectively (see \fref{fig:all_plots}~C and D). 
Moreover, STRidge with $d_{tol}=2$ is more proficient in identifying the quantum harmonic oscillator and the transport equation with an SNR lower than 52 dB, see \fref{fig:all_plots}~E and \ref{fig:all_plots2}~C, respectively. 
These results from the SNR and $N$ experiments reveal that using a single fixed threshold in STRidge can lead to performance variability depending on the input data, highlighting the difficulty of selecting an appropriate $d_{tol}$ threshold without prior knowledge of the system. 
This variability underscores the sensitivity of STRidge to specific threshold settings, which can impact its consistency across different datasets.
Overall, STRidge surpasses ARGOS-RAL in identifying simpler PDEs, such as the transport and diffusion equations; see \fref{fig:all_plots2} C and D.

\subsection{Robustness Analysis using White Gaussian Noise}

To better understand the limits of identification algorithms, we designed an extreme test on a single spatial dimension. This test effectively creates a situation without valid data collection ($\sigma_U=0$), equivalent to an SNR of negative infinity, representing a dataset entirely composed of random noise. 
This scenario sets the ultimate test stage for an algorithm: identifying dynamical systems without signal, where we expect success rates to drop to zero. 
When faced with this condition, an effective algorithm should identify either a null model (with no coefficients) or a dense model (with many terms from the candidate library).
However, if the algorithm incorrectly identifies canonical PDEs from pure white noise data, it indicates that further improvements are needed to prevent such misidentifications and ensure the robustness of the method.

We generate 100 white Gaussian noise datasets, each consisting of 2000 spatial ($x$) and 1000 temporal ($t$) data points, forming a matrix in $\mathbb{R}^{2000\times1000}$.
To investigate the influence of noise variance on the identification process, we use three Gaussian distributions with variances spanning three orders of magnitude: $N(0,0.1^2)$, $N(0,1)$, and $N(0,10^2)$. 
We aim to determine whether ARGOS-RAL and STRidge can identify canonical PDEs under these noise conditions.
Table~\ref{tab:random_test} shows the percentages of different identified models. 
Based on the PDEs tested by Rudy \etal~\cite{rudy_data-driven_2017} and our own study, we define parsimonious models as those having three or fewer nonzero coefficients, suggesting they may correspond to specific physical phenomena. 
In particular, we highlight three classic differential equations: the ODE $u_t = c_1u^d$, the transport equation $u_t = c_2u_x$, and the heat equation $u_t = c_3u_{xx}$.
In contrast, we classify models with more than three nonzero coefficients as non-parsimonious, indicating that their coefficient vectors have a dense composition.

\begin{figure}[htp]
    \centering
    \includegraphics[width=0.9\textwidth]{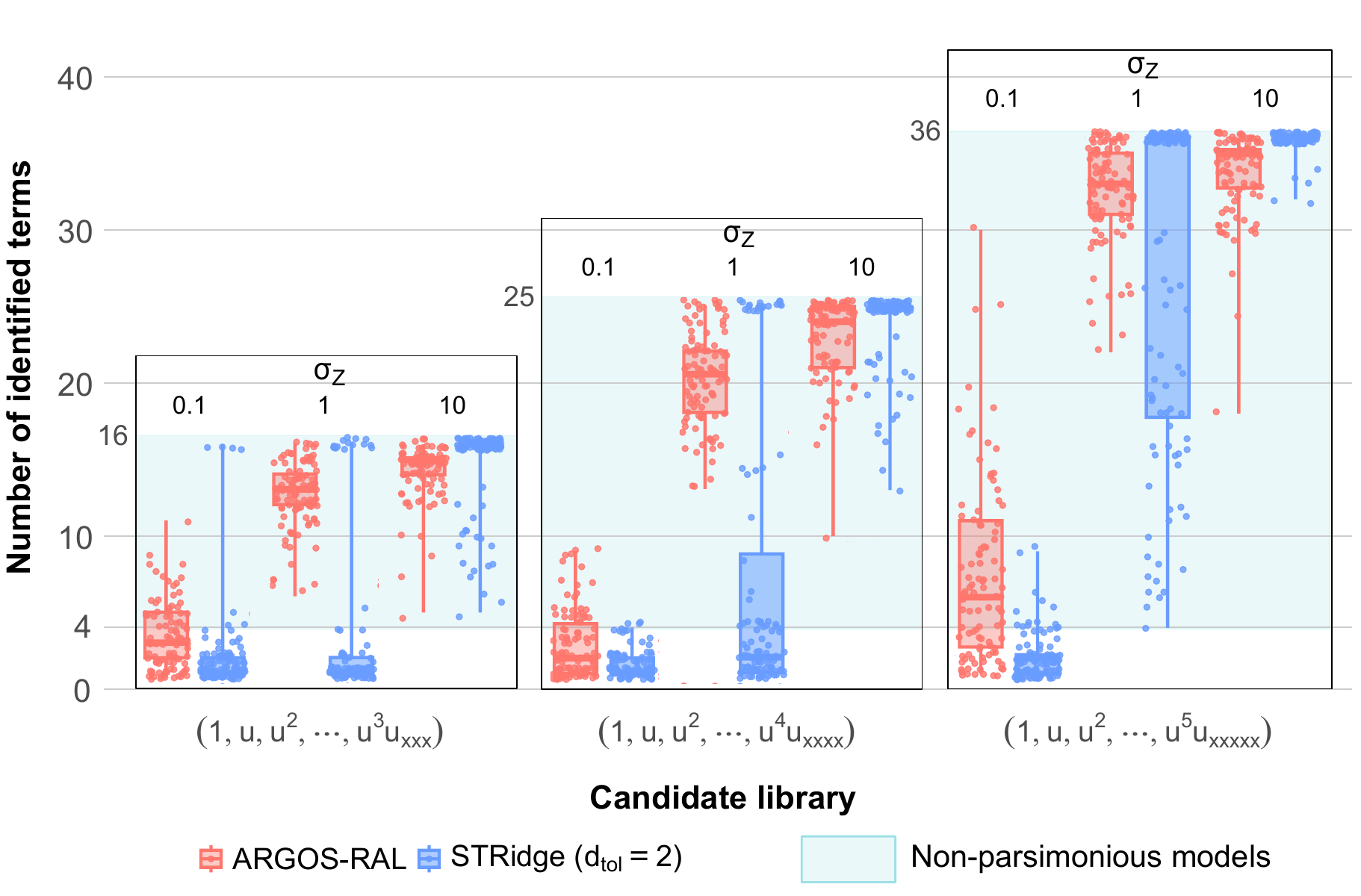}
    \caption{Number of nonzero terms identified from 100 random noise datasets using different candidate function libraries. For each case, we count the number of nonzero coefficients in the sparse regression. We display the distribution of these counts using dots for each of the 100 trials and summarize the results using box plots. Each box plot shows the median (solid horizontal line), interquartile range (box), and minimum and maximum values (whiskers) for the 100 trials. The optimal algorithm should produce boxes located either at zero, indicating a null model, or above four, representing a dense model. The box may span a wide range from four to the maximum number of terms in the library.
    }
    \label{fig:boxplot}
\end{figure}

Table~\ref{tab:random_test} and \fref{fig:boxplot} demonstrate that as the standard deviation of the Gaussian noise increases, both ARGOS-RAL and STRidge tend to identify more non-parsimonious models, as indicated by the probability distributions of the number of identified terms shifting into the shaded region of \fref{fig:boxplot}. This is the desired behavior when the input signal is pure white noise, as we want to ensure that the algorithms do not identify parsimonious models in such cases.
The difference in behavior between the two methods is most apparent when the noise level is low to moderate ($\sigma_Z \leq 1$). In these cases, STRidge's distributions are more spread out and partially located in the parsimonious region, while ARGOS-RAL's distributions are more concentrated in the non-parsimonious region. This suggests that ARGOS-RAL is more effective at avoiding the identification of parsimonious models when the input signal is pure white noise with low to moderate noise levels.
As the noise level increases to $\sigma_Z = 10$, both ARGOS-RAL and STRidge consistently identify non-parsimonious models, as evidenced by the concentration of their distributions in the non-parsimonious region of \fref{fig:boxplot}. This indicates that both methods are effective at avoiding the identification of parsimonious models when the input signal is pure white noise with high noise levels.

\begin{table}[htp]
    \centering
    \begin{tabularx}{\textwidth}{c|*{4}{>{\centering\arraybackslash}X|}c}
    \hline
    \multicolumn{1}{c|}{\multirow{3}{*}{Candidate library}} & \multicolumn{4}{c|}{Parsimonious model (\%)} & \multicolumn{1}{c}{\multirow{2}{*}{\begin{tabular}[c]{@{}c@{}c@{}}Non-\\parsimonious\\ model (\%)\end{tabular}}} \\ 
    \cline{2-5}
    \multicolumn{1}{c|}{} & \multicolumn{1}{c|}{\begin{tabular}[c]{@{}c@{}}ODE\\ $u_t=c_1u^{d}$\end{tabular}} & \multicolumn{1}{c|}{\begin{tabular}[c]{@{}c@{}}Transport\\ $u_t=c_2u_x$\end{tabular}} & \multicolumn{1}{c|}{\begin{tabular}[c]{@{}c@{}}Heat\\ $u_t = c_3u_{xx}$\end{tabular}} & \multicolumn{1}{c|}{Others} & \multicolumn{1}{c}{}  \\ 
    \hline 
    $\sigma_Z = 0.1$ \\
    \hline
    \textbf{STRidge} ($d_{tol}$=2) & & & & & \\
    $(1,u,u^2,\dots,u^3u_{xxx})$ & 4 (1) & 3 & 2 (1) & 82 (11) & 9 \\
    $(1,u,u^2,\dots,u^4u_{xxxx})$ & 0 & 0 & 0 & 91 (10) & 9 (2)\\
    $(1,u,u^2,\dots,u^5u_{xxxxx})$ & 0 & 0 & 0 & 81 (11) & 19 (1)\\
    \hline 
    \textbf{ARGOS-RAL} & & & & & \\
    $(1,u,u^2,\dots,u^3u_{xxx})$ & 1 (1) & 2 (1) & 4 (2) & 54 (29) & 39 (23) \\
    $(1,u,u^2,\dots,u^4u_{xxxx})$ & 0 & 2 (1) & 3 (2) & 63 (32) & 32 (23)\\
    $(1,u,u^2,\dots,u^5u_{xxxxx})$ & 0 & 0 & 0 & 34 (18) & 66 (47)\\
    \hline
    $\sigma_Z = 1$ \\
    \hline
    \textbf{STRidge} ($d_{tol}$=2) & & & & & \\
    $(1,u,u^2,\dots,u^3u_{xxx})$ & 4 (1) & 4 & 4 (2) & 69 (8) & 19 (1) \\
    $(1,u,u^2,\dots,u^4u_{xxxx})$ & 1 & 0 & 2 & 54 (7) & 43 (10)\\
    $(1,u,u^2,\dots,u^5u_{xxxxx})$ & 0 & 0 & 0 & 0 & 100 (18)\\
    \hline 
    \textbf{ARGOS-RAL} & & & & & \\
    $(1,u,u^2,\dots,u^3u_{xxx})$ & 0 & 0 & 0 & 0 & 100 (16) \\
    $(1,u,u^2,\dots,u^4u_{xxxx})$ & 0 & 0 & 0 & 0 & 100 (17)\\
    $(1,u,u^2,\dots,u^5u_{xxxxx})$ & 0 & 0 & 0 & 0 & 100 (13)\\
    \hline
    $\sigma_Z = 10$ \\
    \hline
    \textbf{STRidge} ($d_{tol}$=2) & & & & & \\
    $(1,u,u^2,\dots,u^3u_{xxx})$ & 0 & 0 & 0 & 0 & 100 (11) \\
    $(1,u,u^2,\dots,u^4u_{xxxx})$ & 0 & 0 & 0 & 0 & 100 (4)\\
    $(1,u,u^2,\dots,u^5u_{xxxxx})$ & 0 & 0 & 0 & 0 & 100 (12)\\
    \hline 
    \textbf{ARGOS-RAL} & & & & & \\
    $(1,u,u^2,\dots,u^3u_{xxx})$ & 0 & 0 & 0 & 0 & 100 (12) \\
    $(1,u,u^2,\dots,u^4u_{xxxx})$ & 0 & 0 & 0 & 0 & 100 (9)\\
    $(1,u,u^2,\dots,u^5u_{xxxxx})$ & 0 & 0 & 0 & 0 & 100 (6)\\
    \hline
    \end{tabularx}
    \caption{Models identified from random noise by ARGOS-RAL and STRidge. We construct the candidate library with monomials and derivatives of orders ranging from three to five. We define parsimonious models as having three or fewer nonzero coefficients. We evaluate each identified model with an F-test to determine statistical significance, with the number of significant models (p-value $< 0.05$) noted in parentheses. Numbers outside parentheses indicate the number of models that did not significantly differ from the null hypothesis according to the F-test. $c_1,c_2,c_3$ are constants. For the ordinary differential equation (ODE) models, the monomial order $d$ is a positive integer, with the maximum order corresponding to the highest order in the candidate library.}
    \label{tab:random_test}
\end{table}

\section{Conclusions}
We designed ARGOS-RAL to automatically tune algorithm hyperparameters, enabling the identification of closed forms of PDEs directly from data. ARGOS-RAL offers several advantages over existing PDE identification methods. First, it automates the process of calculating partial derivatives and constructing the candidate library, reducing manual intervention and streamlining the modeling process. Second, the recurrent adaptive lasso employed by ARGOS-RAL provides a more robust and efficient sparse regression technique compared to the STRidge used in SINDy-based methods. This enables ARGOS-RAL to handle noisy and limited data more effectively, as demonstrated in our numerical experiments.

However, ARGOS-RAL also has some limitations. Like other library-based methods, its effectiveness depends on including the correct governing terms in the candidate library. If the true governing terms are absent, ARGOS-RAL can only approximate the PDE using the available terms, potentially leading to model misspecification. Furthermore, while ARGOS-RAL provides a more computationally efficient approach than ARGOS~\cite{egan_automatically_2024} by focusing on point estimates rather than bootstrapping for confidence intervals, this comes at the cost of losing uncertainty quantification for the estimated coefficients.

When applying ARGOS-RAL to different scientific domains, several challenges arise. One key challenge is determining the appropriate range of candidate terms to include in the library, which often requires domain expertise. In some fields, the governing equations may involve complex nonlinearities or unconventional terms that are difficult to anticipate without prior knowledge. Another challenge is the computational cost of handling high-dimensional data, which is common in many scientific applications. As the number of variables and the complexity of the PDE increase, the size of the candidate library grows exponentially, leading to increased computational demands for sparse regression.

Despite these challenges, ARGOS-RAL offers a promising framework for automating PDE identification in various scientific domains. By leveraging sparse regression techniques and automating key steps in the modeling process, ARGOS-RAL has the potential to accelerate discovery and insight in fields ranging from physics and engineering to biology and climate science.

\section*{Data availability}
All data and codes are available at \url{https://github.com/Weizhenli/ARGOS-RAL}.

\clearpage

\clearpage
\appendix
\section{Supplementary materials}

\subsection{Gaussian Blur Kernels} \label{app:GB}
The Gaussian blur convolves data with a Gaussian kernel to smooth it, regardless of the data's dimensionality. This convolution method offers significant benefits for filtering out Gaussian noise, a common noise distribution encountered in data analysis~\cite{gonzalez_smoothing_2018}.
For one-dimensional spatial PDEs, such as the Burgers' and cable equations, we employ the simplest 2-dimensional Gaussian kernel:
\begin{equation}\label{eq:kernel2} 
    \frac{1}{16} \left[\begin{array}{c} 1 \\ 2 \\ 1 \end{array}\right] \otimes \left[\begin{array}{ccc} 1 & 2 & 1 \end{array}\right] = \frac{1}{16} \left[\begin{array}{ccc} 1 & 2 & 1 \\ 2 & 4 & 2 \\ 1 & 2 & 1 \end{array}\right].
\end{equation}
In contrast, for two-dimensional spatial PDEs, the Navier-Stokes and reaction-diffusion equations, we use a 3-dimensional Gaussian kernel: 
\begin{equation}\label{eq:kernel3} 
    \frac{1}{64} \left[\begin{array}{c} 1 \\ 2 \\ 1 \end{array}\right] \otimes \left[\begin{array}{ccc} 1 & 2 & 1 \end{array}\right] \otimes \left[\begin{array}{ccc} 1 \\ 2 \\ 1 \end{array}\right] 
    = \frac{1}{64} \left[\left[\begin{array}{ccc} 1 & 2 & 1 \\ 2 & 4 & 2 \\ 1 & 2 & 1 \end{array}\right] \left[\begin{array}{ccc} 2 & 4 & 2 \\ 4 & 8 & 4 \\ 2 & 4 & 2 \end{array}\right] \left[\begin{array}{ccc} 1 & 2 & 1 \\ 2 & 4 & 2 \\ 1 & 2 & 1 \end{array}\right]\right].
\end{equation}

\subsection{Algorithms} \label{app:algos}
Here, we detail the automatic Savitzky-Golay Filter and the recurrent adaptive lasso with the Pareto curve and AIC.
\begin{algorithm}[htp] 
\caption{Automatic Savitzky-Golay Filter}
\label{al:SG} 
\LinesNumbered
\KwIn{$\mathbf{U} \in \mathbb{R}^{n\times m}$ or $\mathbb{C}^{n\times m}$, $dt$, $dx$.}
\KwOut{partial derivatives $\mathbf{U}_t$, $\mathbf{U}_x$, $\mathbf{U}_{xx}$, $\cdots$.}

$\mathbf{U}_\mathrm{GB} = $ Gaussian\_Blur$(\mathbf{U})$; \tcp{ use Gaussian blurred data as the ground truth; } 
$(o^{*}_t,l^{*}_t) = \underset{o,l}{\arg\min}$ MSE (Savitzky-Golay$(\tilde{\mathbf{U}}(t),o,l)$, $\mathbf{U}_\mathrm{GB}$)\;
$(o^{*}_x,l^{*}_x) = \underset{o,l}{\arg\min}$ MSE (Savitzky-Golay$(\tilde{\mathbf{U}}(x),o,l)$, $\mathbf{U}_\mathrm{GB}$)\;

$\mathbf{U}_t =$ Savitzky-Golay($\mathbf{U}_{GB}, o^{*}_t, l^{*}_t,$  derivative=1)\; 
$\mathbf{U}_x =$ Savitzky-Golay($\mathbf{U}_{GB}, o^{*}_x, l^{*}_x,$  derivative=1)\; 
$\mathbf{U}_{xx} =$ Savitzky-Golay($\mathbf{U}_{GB}, o^{*}_x, l^{*}_x,$  derivative=2)\;
$\vdots$
\\
\end{algorithm}

\SetKwComment{Comment}{/*}{*/}
\begin{algorithm}[!htp] 

\caption{The recurrent adaptive lasso with Pareto curve and AIC}
\label{al:reada} 
\KwIn{$\Theta(u) \in \mathbb{R}^{(n\cdot m)\times p}$ or $\mathbb{C}^{(n\cdot m)\times p}$, $u_t \in \mathbb{R}^{(n\cdot m)\times 1}$ or $\mathbb{C}^{(n\cdot m)\times 1}$.}
\KwOut{$\hat{\beta}$}

\nl \For{$\gamma$ in 1:5}{
    \nl $\mathcal{J}^{(\gamma,0)} = \mathrm{NULL}$; \tcp{ initialize $\mathcal{J}$;}
    \nl k = 1; \tcp{iteration counter;}
    \nl $\mathcal{J}^{(\gamma,k)} = \left\{1,2,\cdots, p\right\}$; \tcp{ selected columns from $\Theta$;}
    \nl \While{$\mathcal{J}^{(\gamma,k)}\neq \mathcal{J}^{(\gamma,k-1)}$}{
        \nl 
        $
            w^{(\gamma,k)} = \left(\mathop{\arg\min}_{\beta_{\mathcal{J}^{(\gamma,k)}}} \left\|u_t - \Theta(u)_{\mathcal{J}^{(\gamma,k)}}\beta_{\mathcal{J}^{(\gamma,k)}}\right\|_2^2\right)^{-\gamma}
        $;  \tcp{ ols weights; }
        \nl 
        $
            \hat{\beta}^{(\gamma,k)} = \mathop{\arg\min}_{\beta_{\mathcal{J}^{(\gamma,k)}}} \left\|u_t -\Theta(u)_{\mathcal{J}^{(\gamma,k)}}\beta_{\mathcal{J}^{(\gamma,k)}}\right\|_2^2 + \lambda^* \sum_{j=1}^pw^{(\gamma,k)}_j|\beta_{j\cdot \mathcal{J}^{(\gamma,k)}}|
        $; \\
        \tcp{ $\lambda^*$ is the optimal point on the Pareto curve; }
        \nl $\mathcal{A}^{(\gamma,k)}$ = AIC($\hat{\beta}^{(\gamma,k)}$)\;
        \nl $\mathcal{J}^{(\gamma,k)} = \left\{j: \hat{\beta}_j^{(\gamma,k)} \neq 0 \right\}$; \tcp{select active terms;}
        \nl $k = k + 1$\;
    }
}

\nl $\mathcal{J}^*= \mathcal{J}^{(\gamma^*, k^*)}$ where $(\gamma^*, k^*)$ is the index of the minimum $\mathcal{A}$\;
\nl $\hat{\beta} = \mathop{\arg\min}_{\beta_{\mathcal{J}^*}} \left\|u_t - \Theta(u)_{\mathcal{J}^*}\beta_{\mathcal{J}^*}\right\|_2^2$\;

\end{algorithm}

\subsection{Additional PDE Test Cases} \label{app:more_eq}
Here, we demonstrate how to solve the PDEs presented in this paper. 

\subsubsection{Burgers' equation}
We can derive Burgers' equation from the Navier-Stokes equation for the velocity field by dropping the pressure gradient term.
Unlike the Navier-Stokes equation, Burgers' equation does not exhibit turbulent behavior, and we can transform it to linear form via the Cole-Hopf transformation~\cite{cross_pattern_1993}:
\begin{equation} \label{eq:bur}
    u_t = -uu_{x} + \nu u_{xx}.
\end{equation}
We solve Burgers' Eq.~\eqref{eq:bur} using the Fourier spectral method~\cite{brunton_fourier_2022} with the \textit{ode45} function in MATLAB. 
We set $\nu=0.1$, $x\in [-8,8]$ with 256 points, $t\in [0,10]$ with 101 points, and the initial condition is a Gaussian function: $\exp\left(-(x+2)^2\right)$. 

\subsubsection{Cable equation}
The cable equation, shown in Eq.~\eqref{eq:cable}, quantitatively describes the electrical behavior of nerve axons and other cable-like structures in biological systems. It captures the electrical circuit of current flow and voltage change both within and between neurons. 
The equation is derived from a circuit model of the membrane and its intracellular and extracellular space.
The cable equation plays a crucial role as an important PDE in biophysical studies, helping researchers understand how electrical signals change in diseases and disorders.
By identifying the cable equation, researchers can diagnose these negative conditions by checking for changes in capacitances $c_{m}$, resistances $r_m$, and axial resistance $r_a, r_e$:
\begin{equation} \label{eq:cable}
    {\lambda}^2\frac{\partial^2V}{\partial {x}^2}=\tau \frac{\partial V}{\partial t}+V\kern1em \mathrm{where}\kern1em \lambda =\sqrt{\frac{r_m}{r_e+{r}_a}}\kern1em \mathrm{and}\kern1em \tau ={r}_m{c}_m.
\end{equation}
We solve the cable equation using \textit{odeint} function in Python with the Fourier spectral method. 
We set $\lambda=1$, $\tau = 1$, $x \in [-4,4]$ with $\Delta x =0.1$, $t \in[0,5]$ with $\Delta t=0.01$, and use a Gaussian function $\exp\left(-x^2\right)$ as the initial condition.

\subsubsection{Navier-Stokes}
We simulate the two-dimensional Navier-Stokes equation for fluid flow around a circular cylinder using the Immersed Boundary Projection Method~\cite{taira_immersed_2007,colonius_fast_2008}.
The two-dimensional velocity components are denoted by $u$ and $v$, while $\omega$ represents the vorticity away from the circular cylinder of diameter one and mass centre at $(x=1,y=2)$.
We set the Reynolds number to 100 and aim to identify the equation
\begin{equation}\label{eq:NS_eq}
    \omega_t=0.01\omega_{xx}+0.01\omega_{yy}-u\omega_x-v\omega_y. 
\end{equation}
The spatial domain spans $x\in[0,9]$ with $\Delta x=0.02$, $y\in[0,4]$ with $\Delta y=0.02$, and the temporal domain covers $t\in[300,330]$ with $\Delta t=0.02$. We save the flow data every ten snapshots.
This setup generates a simulated dataset containing approximately 13.5 million points ($449 \times 199 \times 151$).
However, constructing the candidate library $\mathbf{\Theta}$ for such a large dataset poses computational challenges.
To facilitate the evaluation, we randomly sample points within the red rectangular area shown in Fig. 4 C at each snapshot.

\subsubsection{Reaction-diffusion}
Reaction-diffusion systems offer a versatile framework to model pattern formation in various natural phenomena in chemistry, biology, geology, physics, and ecology. These systems give rise to a rich tapestry of periodic patterns, including spots, zigzags, spiral waves, and rolls. In our analysis, we focus on a widely studied class of reaction-diffusion systems known as the $\lambda-\omega$ systems, described by the following coupled PDEs:
\begin{eqnarray} \label{eq:RD}
    u_t &= 0.1u_{xx}+0.1u_{yy}+u-uv^2-u^3+v^3+u^2v  \\ 
    v_t &= 0.1v_{xx}+0.1v_{yy}+v-uv^2-u^3-v^3-u^2v 
\end{eqnarray}
To generate data for our analysis, we employ the simulation method described in~\cite{rudy_data-driven_2017}. We discretize the spatial domain $x,y \in [-10,10]$ using a $512\times512$ grid and evolve the system over the time interval $t\in [0,10]$ using 201 time steps. This procedure yields a rich dataset comprising 52,690,944 spatiotemporal points on a $512\times512\times201$ grid, providing a comprehensive characterization of the system's dynamics.

\subsubsection{Quantum harmonic oscillator}
The quantum harmonic oscillator (QHO) models the parabolic potential of a harmonic oscillator in quantum mechanics. It simulates the time evolution of the wave function associated with a particle in the parabolic potential, providing the probability distribution of the particle's position at any given time by taking the squared magnitude of the wave function. The energy levels of a quantum harmonic oscillator are quantized, meaning they can only assume specific, discrete values. Furthermore, even if we form a statistical distribution from multiple experiments, it will lack information on the intricate phase of the wave function. We use the following equation:
\begin{equation} \label{eq:qho}
    u_t = \frac{1}{2}iu_{xx}-iuV  = \frac{1}{2}iu_{xx}-\frac{x^2}{2}iu. 
\end{equation}
To obtain data on the QHO, we employ the operator splitting method with the Fourier transform. We consider the time domain $t\in [0,10]$ with $\Delta t=0.025$, and the space domain $x\in [-7.5,7.5]$ with $\Delta x=15/512$, using a Gaussian $\exp(-((x-1)/2)^2)$ as the initial condition. 
When performing a sparse regression on complex numbers, we transform the regression from complex to real numbers. 
For each $y_i = y_i^R+iy_i^I$, where the normal $i$ is the imaginary number, the subscript $_i$ represents the $i$th observation, we can reform it as
\begin{align*}
    &y_i^R+iy_i^I =  \beta_0^R+i\beta_0^I + \sum_{j=1}^p\left[(\beta_j^R+i\beta_j^I)(x_{ij}^R+ix_{ij}^I)\right]+\epsilon^R+i\epsilon^I\\
    &=\beta_0^R+i\beta_0^I+(\beta_1^R+i\beta_1^I)(x_{i1}^R+ix_{i1}^I)+ (\beta_2^R+i\beta_2^I)(x_{i2}^R+ix_{i2}^I)+\cdots + \\
    & \qquad (\beta_p^R+i\beta_p^I)(x_{ip}^R+ix_{ip}^I)+ \epsilon^R+i\epsilon^I \\
    \vspace{5pt}
    &=\beta_0^R+i\beta_0^I+\sum_{j=1}^p(x_{ij}^R\beta_j^R-x_{ij}^I\beta_j^I) + 
    i\sum_{j=1}^p(x_{ij}^R\beta_j^I+x_{ij}^I\beta_j^R) +\epsilon^R+i\epsilon^I 
\end{align*} 
and split it to extract two equations for both real and imaginary parts
\begin{equation} \label{eq:complex}
    y_i^R = \beta_0^R+\sum_{j=1}^px_{ij}^R\beta_j^R-\sum_{j=1}^px_{ij}^I\beta_j^I+\epsilon^R, \\
    y_i^I = \beta_0^I+\sum_{j=1}^px_{ij}^I\beta_j^R+\sum_{j=1}^px_{ij}^R\beta_j^I+\epsilon^I.
\end{equation} \par
Based on Eq.~\eqref{eq:complex}, we can organize the dataset as:
\begin{equation*}
        Y=\left[\begin{array}{c}
y_1^R\\y_1^I\\y_2^R\\y_2^I\\\vdots\\y_n^R\\y_n^I
\end{array}\right], \hspace{8pt}
\mathbf{X}=\left[\begin{array}{ccccccc}
    x_{11}^R & -x_{11}^I & x_{12}^R & -x_{12}^I & \cdots & x_{1p}^R & -x_{1p}^I\\
    x_{11}^I & x_{11}^R & x_{12}^I & x_{1R}^R & \cdots & x_{1p}^I & x_{1p}^R\\
    x_{21}^R & -x_{21}^I & x_{22}^R & -x_{22}^I & \cdots & x_{21}^R & -x_{21}^I\\
    x_{21}^I & x_{21}^R & x_{22}^I & x_{22}^R & \cdots & x_{21}^I & x_{21}^R\\
    \vdots & \vdots & \vdots & \vdots & \ddots & \vdots & \vdots \\
    x_{n1}^R & -x_{n1}^I & x_{n2}^R & -x_{n2}^I & \cdots & x_{n1}^R & -x_{n1}^I \\
    x_{n1}^I & x_{n1}^R & x_{n2}^I & x_{n2}^R & \cdots & x_{n1}^I & x_{n1}^R
\end{array}\right].
\end{equation*}
Finally, we implement the recurrent adaptive lasso and STRidge on the re-posited $Y$ and $\mathbf{X}$.

\subsubsection{Advection-diffusion equation}
The advection-diffusion equation, which combines advection and diffusion terms, describes the transport and dispersion of quantities such as temperature, substance concentration, or fluid velocity in various scientific and engineering contexts. We can express this equation as follows:
\begin{equation}
    c_t = Dc_{xx} - uc_x
\end{equation}
We solve the advection-diffusion equation using the Fourier spectral method and the \textit{odeint} function in Python. We set the diffusion coefficient $D=1$, the advection velocity $u=1$, and consider the spatial domain $x \in [-10,10]$ with resolution $\Delta x =0.1$ and the temporal domain $t \in[0,10]$ with resolution $\Delta t=0.01$. We use a Gaussian function of the form $\exp\left(-(x+2)^2\right)$ as an initial condition.

\subsubsection{The KdV equation}
The Korteweg–De Vries (KdV) equation describes wave propagation on shallow water surfaces. The KdV equation solution reveals that an isolated traveling wave exhibits linear behavior, but nonlinear interactions emerge when multiple waves are present. Moreover, the dependence of wave velocity on wave amplitude ensures that any solution with multiple amplitudes will display nonlinear behavior, regardless of the interaction. Eq.~\eqref{eq:kdv} presents the formula for the KdV equation:
\begin{equation} \label{eq:kdv}
    u_t=-6uu_x-u_{xxx}.
\end{equation} 
We employ the two-soliton solution~\cite{polyanin_third-order_2012} to solve the KdV equation: 
\begin{align}
    w(x, t)=-2 \frac{\partial^2}{\partial x^2} \ln \left(1+B_1 e^{\theta_1}+B_2 e^{\theta_2}+A B_1 B_2 e^{\theta_1+\theta_2}\right) \label{eq:sol_kdv} \\
    \theta_1=a_1 x-a_1^3 t, \quad \theta_2=a_2 x-a_2^3 t, \quad A=\left(\frac{a_1-a_2}{a_1+a_2}\right)^2 \nonumber
\end{align} 
where $a_1$, $a_2$, $B_1$, and $B_2$ are arbitrary constants.
We set the following parameters: 201 time steps ($n=201$) with $t\in [0,20]$, 512 spatial points ($m=512$) with $x\in [-30,-30]$, and $a_1=0.5$, $a_2=1$, $B_1=1$, $B_2=5$.

\subsubsection{Transport equation}
The transport equation, Eq.~\eqref{eq:trans}, plays a fundamental role in science and engineering, describing the spatiotemporal evolution of scalar quantities or vector fields. We solve this PDE using the analytical solution, Eq.~\eqref{eq:trans2}, with $c=3$ to generate the data for our study. The spatial domain spans $x \in [-5,1]$ with resolution $\Delta x=0.01$, while the temporal domain covers $t \in[0,2]$ with timestep $\Delta t = 0.01$.
\begin{eqnarray}
    &u_t = cu_{x}, \quad c>0 \label{eq:trans} \\
    &u(x,t) = \exp(-(x+ct)^2) \label{eq:trans2}
\end{eqnarray}

\subsubsection{Diffusion equation}
The diffusion (heat) equation, Eq.~\eqref{eq:heat}, plays a crucial role in many scientific and engineering fields, including solid-state physics, materials science, environmental science, and computational fluid dynamics. This equation elucidates the fundamental process of heat diffusion, enabling engineers to gain deep insights into heat conduction, thermal conductivity, and temperature-dependent phenomena in solids and other materials.
Here, we use the analytic solution, Eq.~\eqref{eq:heat_sol}, to generate the data. We set $x \in [0,5]$ with $\Delta x=0.01$ and $t \in[0,1.5]$ with $\Delta t = 0.01$, and choose the initial condition as $6\sin \left( \pi x / L \right)$.
\begin{equation} \label{eq:heat}
    u_t = 10u_{xx}
\end{equation} 
\begin{equation} \label{eq:heat_sol}
    u\left( {x,t} \right) = 6\sin \left( {\frac{{\pi x}}{L}} \right){{\bf{e}}^{ - k{{\left( {\frac{\pi }{L}} \right)}^2}\,t}}, \quad k=10
\end{equation} 
The diffusion equation and its analytic solution provide a powerful framework for understanding and predicting heat transfer in various systems. By carefully selecting the spatial and temporal domains and the initial condition, we can model a wide range of real-world scenarios and gain valuable insights into the underlying physical processes.

\begin{figure}[htp]
    \centering
    \includegraphics[width=1\textwidth]{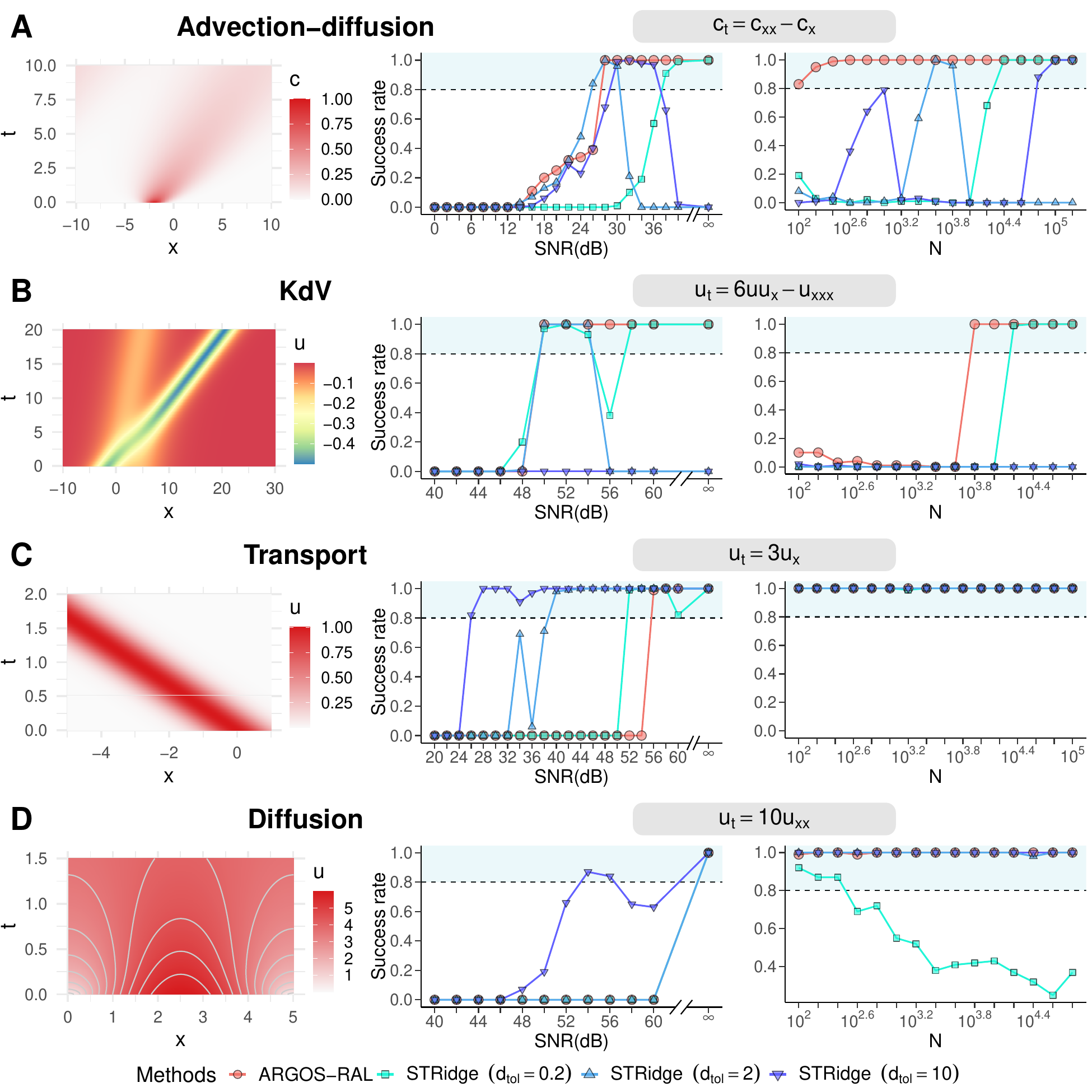}
    \caption{Success rates of ARGOS-RAL and STRidge in identifying (A) advection-diffusion, (B) KdV, (C) transport, and (D) heat equations with different SNRs and sample sizes. 
    To analyze the noise tolerance, we added noise to the PDE solutions at different SNR levels.
    For the sample size analysis, we randomly sampled points from the set $\{\mathbf{u}_t,\mathbf{\Theta}(\mathbf{u})\}$ based on noiseless PDE solutions. 
    For sample size analysis, we randomly sample points from $\{\mathbf{u}_t,\mathbf{\Theta}(\mathbf{u})\}$ set based on noiseless PDE solutions. Lines are only used to link points, not to fit points. The plots demonstrate that our method maintains high success rates in identifying the correct PDE even under significant noise and with limited sample sizes.
    }
    \label{fig:all_plots2}
\end{figure}

\end{document}